# Densely Multiplied Physics Informed Neural Networks

Feilong Jiang[a]    Xiaonan Hou[a]*   Min Xia[b]*
[a] Department of Engineering, Lancaster University, LA1 4YW Lancaster, U.K.
[b] Department of Mechanical and Materials Engineering, University of Western Ontario, London, Ontario, Canada
x.hou2@lancaster.ac.uk
mxia47@uwo.ca

**Abstract**

Although physics-informed neural networks (PINNs) have shown great potential in dealing with nonlinear partial differential equations (PDEs), it is common that PINNs will suffer from the problem of insufficient precision or obtaining incorrect outcomes. Unlike most of the existing solutions trying to enhance the ability of PINN by optimizing the training process, this paper improved the neural network architecture to improve the performance of PINN. We propose a densely multiply PINN (DM-PINN) architecture, which multiplies the output of a hidden layer with the outputs of all the behind hidden layers. Without introducing more trainable parameters, this effective mechanism can significantly improve the accuracy of PINNs. The proposed architecture is evaluated on four benchmark examples (Allan-Cahn equation, Helmholtz equation, Burgers' equation and 1D convection equation). Comparisons between the proposed architecture and different PINN structures demonstrate the superior performance of the DM-PINN in both accuracy and efficiency.

**Keywords:** Physics-informed neural networks, Scientific machine learning, Numerical methods for PDE

## 1. Introduction

Physics-informed neural networks (PINNs) [1] have been proven to be an effective machine learning model for solving partial different equations (PDEs). By imposing the PDE residuals and boundary conditions into the loss function of the neural network, the PINNs could be trained with little or no labelled data [2-4]. Utilizing this strategy, the output of a trained PINN can satisfy the given physical law. Therefore, PINN-based approaches have better generalization capability than data-driven machine learning methods, especially when the data is limited. Even though PINN method has been successfully applied to a variety of science and industrial fields [5-9], it can still suffer from low precision or even failure in solving the PDE [10, 11]. Many efforts have been made by researchers to improve the accuracy of PINN. One possible failure mode of the PINN is the imbalanced residual distribution [12]. During the process of calculating the loss value of the neural network, a mean value of all collocation points is chosen. Under some circumstances, points that hold large residuals may only occupy a small part of the whole training data set. Consequently, the importance of these points will be neglected during the training process. These points will not be properly handled, which leads to the wrong solution. To overcome this problem, Daw et al. proposed an evolutionary sampling algorithm [12]. In this algorithm, the training points will be chosen randomly in every training iteration. The points with the residual higher than the criterion will be held to the next iteration. With this sampling strategy, the PINN achieved better performance than the traditional one. Similar work was proposed by Wu et al. [13] but the resampling method varies. Another method for resolving this problem is to assign higher weights to those high residual points. Unlike the resampling method, the training points of this method are fixed during the training process. A weight matrix was created to apply different weight values to different training points. McClenny et al. proposed the self-adaptive physics-informed neural networks [14]. The weight matrix will be updated according to the derivative of the weights with respect to the loss. Thus, the neural network will cast more attention to the regions that are hard to resolve. Anagnostopoulos conducted the idea in a more direct way [15]. The weight matrix is updated by the normalized loss value of each point and no extra derivative operation is needed.

Another important reason causing the failure of PINN is the spatial-temporal causal problem. Unlike traditional numerical calculation method, the principle of causality is absent in the training process of PINNs. Because all the time and spatial points are inputted into the neural network at the same time, and there is no guarantee that the solutions of former times will be resolved before later times. To integrate the causality into PINN, Wang et al. re-formulated the loss function of PINN [16]. For the re-formulation, only when the residual of the previous time is small enough will the later time proceed. Krishnapriyan et al. dealt with this problem from the aspect of training strategy [10]. Instead of using the entire space-time points for training together, they chose to divide the whole domain into several smaller sequences. For each training process, the model learns to predict only one-time sequence and the end of the sequence is used as the initial state of the next sequence. Thus, the marching-in-time scheme could be applied to the whole training process.

Applying the auxiliary method to the fully connected neural network (FC-NN) itself is also an effective way to improve the performance of PINN. Normally, the boundary conditions will be integrated into the PINN as the loss terms. Sukumar et al. proposed the method of exact imposition of boundary conditions in PINN [17]. In this method, the outputs of PINN satisfy the boundary conditions automatically. Consequently, the PINN only needs to deal with the governing equation of the interior collocation points, which helps to improve the accuracy of PINN. To increase the performance of PINN, Jagtap introduced a trainable parameter into the activation function [18]. Through this method, the convergence speed and accuracy of the neural network increase.

Because improving the structure is also a common method to enhance the ability of a deep learning model [19-23] and the original PINN simply utilized the FC-NN [1]. Adopting a more complex structure can be feasible to improve the expressive power of PINNs. However, only a few researchers have tried to enhance PINNs in this aspect.

Wang et al. successfully improved the accuracy of PINN by introducing the residual connections and multiplicative interactions method to the structure of FC-NN [24]. In this method, two extra single layer networks are needed to project the input variables to a high-dimensional feature space and these features are used to update the hidden layers. To improve the training efficiency and performance, Miao et al., Du et al. and Cheng et al. all tried to introduce the ResNet structure into PINN [25-27]. The adoption of the ResNet structure helps to solve the gradient vanishing problem and unify the linear and nonlinear coefficients.

In this work, we proposed a densely multiply neural network structure to improve the expressive ability of PINNs. We demonstrated the effectiveness of our method with some benchmark PDE problems. The outcomes showed that our method could improve the accuracy of PINN without introducing extra trainable parameters.

The paper is structured as follows. In section 2, we give a brief introduction to Physics-informed neural networks (PINNs) and their background. In section 3, we describe the proposed methodology in detail, and we also introduce the setting of the training process. Section 4 presents the results of numerical examples and the comparison between other neural network structures. In section 5, we discuss the performance of the proposed method from the perspectives of gradient flow dynamics and efficiency. Finally, section 6 is the conclusion of our work.

## 2. Physics-informed neural networks

The physics-informed neural network algorithm is designed to deal with the partial differential equations (PDEs) in a general form of:

$$u_t + \mathcal{N}[u] = 0, \ x \in \Omega, \ t \in [0, \ T], \tag{1}$$

with the initial condition expressed as:

$$u(0, \ x) = h(x), \ x \in \Omega, \tag{2}$$

and boundary conditions shown below:

$$u(t, \ x) = g(t, \ x), \ x \in \partial\Omega, \ t \in [0, \ T], \tag{3}$$

where $u(t, \ x)$ is the latent solution, $\mathcal{N}[\cdot]$ denotes the nonlinear operator parameterized by $\lambda$, $\Omega$ is the spatial domain, $\partial\Omega$ is the boundary of the spatial domain, $x$ and $t$ represent the space and time coordinates, $h(x)$ denotes the initial condition and $g(t, \ x)$ represents the boundary conditions.

The core idea of PINN is to regard the output of the neural network as the solution of PDE and integrate the PDE, initial condition and boundary conditions into the loss function of the neural network. Under this circumstance, the output of the neural network can be written in the form of $u_\theta(t, \ x)$, $\theta$ denotes the trainable parameters of the neural network such as weights and biases. The loss function of the neural network

is shown below:

$$\mathcal{L}(\theta) = \lambda_{ic}\mathcal{L}_{ic}(\theta) + \lambda_{bc}\mathcal{L}_{bc}(\theta) + \lambda_{r}\mathcal{L}_{r}(\theta), \tag{4}$$

where $\mathcal{L}_{ic}(\theta)$ represents the loss term corresponding to the initial condition, $\mathcal{L}_{bc}(\theta)$ denotes the loss term of boundary conditions and $\mathcal{L}_{r}(\theta)$ is the loss term of the PDE. The specific forms of these loss terms are shown below:

$$\mathcal{L}_{ic}(\theta) = \frac{1}{N_{ic}}\sum_{i=1}^{N_{ic}}\left|u_{\theta}\left(0, x_{ic}^{i}\right) - h\left(x_{ic}^{i}\right)\right|^{2}, \tag{5}$$

$$\mathcal{L}_{bc}(\theta) = \frac{1}{N_{bc}}\sum_{i=1}^{N_{bc}}\left|u_{\theta}\left(t_{bc}^{i}, x_{bc}^{i}\right) - g\left(t_{bc}^{i}, x_{bc}^{i}\right)\right|^{2}, \tag{6}$$

$$\mathcal{L}_{r}(\theta) = \frac{1}{N_{r}}\sum_{i=1}^{N_{r}}\left|\frac{\partial u_{\theta}}{\partial t}\left(t_{r}^{i}, x_{r}^{i}\right) + \mathcal{N}[u_{\theta}]\left(t_{r}^{i}, x_{r}^{i}\right)\right|^{2}, \tag{7}$$

where $\left\{x_{ic}^{i}\right\}_{i=1}^{N_{ic}}$ denotes the initial data, $\left\{t_{bc}^{i}, x_{bc}^{i}\right\}_{i=1}^{N_{bc}}$ represents the boundary data and $\left\{t_{r}^{i}, x_{r}^{i}\right\}_{i=1}^{N_{r}}$ denotes the collocation points spreading in the whole space-time domain. The hyper-parameters $\lambda_{ic}$, $\lambda_{bc}$, $\lambda_{r}$ are the weight coefficients, which are used for solving the imbalance problem between each loss term [24, 28].

The training process of PINN is to adjust the trainable parameters $\theta$ to minimize the loss function. Thus, the training of PINN is a process of solving the optimisation problem:

$$\bar{\theta} = \arg\min_{\theta}\mathcal{L}(\theta). \tag{8}$$

The gradient descent algorithm such as Adam [29] and SGD [30] are usually used to implement the minimizing process.

## 3. Methodology

Many studies show that the fully-connected neural network structure can successfully deal with complex PDEs [6, 31-33], and even a single hidden neural network is able to accurately approximate any nonlinear continuous functionals and nonlinear operators [34]. However, the PINN method still suffers from the problem of low precision or even failure in obtaining the correct solution [10, 16]. As an effective method, improving the structure of the base model showed great success in the fields of machine version [19, 20, 35] and natural language processing [21, 36]. Inspired by the DenseNet structure, we introduce a densely multiplied PINN (DM-PINN) structure. With the reusage of the previous outputs, the structure could improve expressive ability without increasing the number of trainable parameters. As shown in Fig. 1, the output of a hidden layer will make element-wise multiplication with each output of all the hidden layers behind.

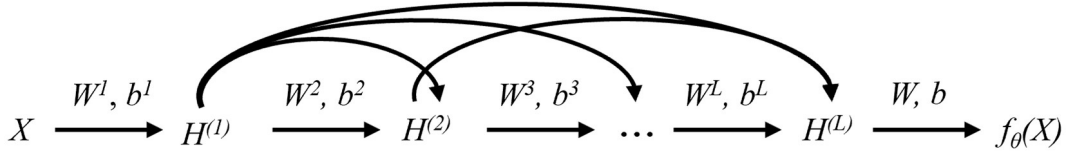

Figure 1: The architectures of densely multiplied neural network.

The output of the hidden layer can be expressed as:

$$H^{(k+1)} = \phi(W^{k+1}H^{(k)} + b^{k+1})\prod_{i=1}^{k} H^{(i)}, \; k = 1,\cdots,L-1, \tag{9}$$

where $H^{(i)}$ denotes the output of the $i^{th}$ hidden layer, $\phi$ denotes the activation function, $W^k$ and $b^k$ represent the weight matrix and bias matrix of the $i^{th}$ hidden layer.

Due to the densely multiplied structure, our method might suffer from the gradient vanishing problem when the number of hidden layer increases. To deal with this problem, we introduced a Batch Normalization layer after the input layer. The ResNet structure [37] as shown in Fig.2 was adopted, and each layer performs as a residual block, which achieves the best performance. Besides, the densely multiplied operation is modified to be a skip multiplied operation as shown in Fig.3. We name this structure as skip densely multiplied PINN (SDM-PINN).

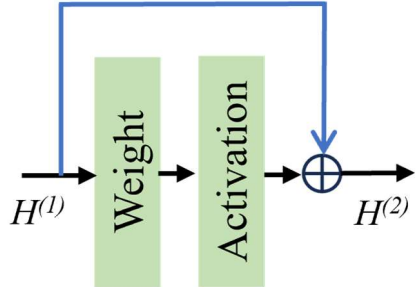

Figure 2: The adopted ResNet structure

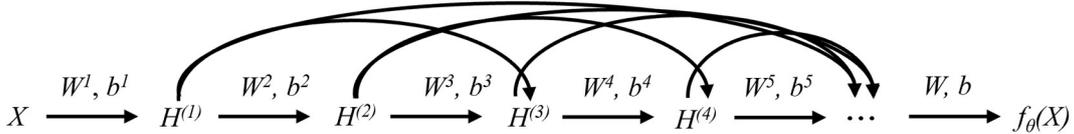

Figure 3: Skip densely multiplied structure

The training data points of all the cases in this work are sampled using Latin Hypercube method. For all the cases, the weights are initialized with Xavier normal distribution, and tanh is used as the activation function. The training is conducted on a single NVIDIA GeForce RTX 4090 GPU. The accuracies of all the cases are averaged over 5 independent trials of each model.

## 4. Numerical examples

## 4.1. Allan-Cahn equation

The first numerical example is the Allan-Cahn equation. This equation is used to model the phase separation process in binary alloys. The equation is expressed as below:

$$u_t - 0.0001 u_{xx} + 5u^3 - 5u = 0, \; x \in [-1, 1], \; t \in [0, 1], \tag{10}$$

$$u(0, x) = x^2 \cos(\pi x), \tag{11}$$

$$u(t, -1) = u(t, 1), \tag{12}$$

$$u_x(t, -1) = u_x(t, 1). \tag{13}$$

To make it easier to implement, we combine the two boundary conditions into one loss term, which could be expressed as:

$$\mathcal{L}_{bc}(\theta) = \frac{1}{N_{bc}} \sum_{i=1}^{N_{bc}} \left( \left| u(t_{bc}^i, 1) - u(t_{bc}^i, -1) \right|^2 + \left| u_x(t_{bc}^i, 1) - u_x(t_{bc}^i, -1) \right|^2 \right). \tag{14}$$

The neural network architecture contains 4 hidden layers with 128 neurons per layer. The number of training data points $N_r$, $N_{ic}$ and $N_{bc}$ are 20000, 100 and 200 (100 points for each boundary respectively). The weight coefficients $\lambda_r$, $\lambda_{ic}$ and $\lambda_{bc}$ are set as 1, 100, and 1 respectively. The model is trained for 15000 iterations with Adam optimizer, and the learning rate is set as 0.001.

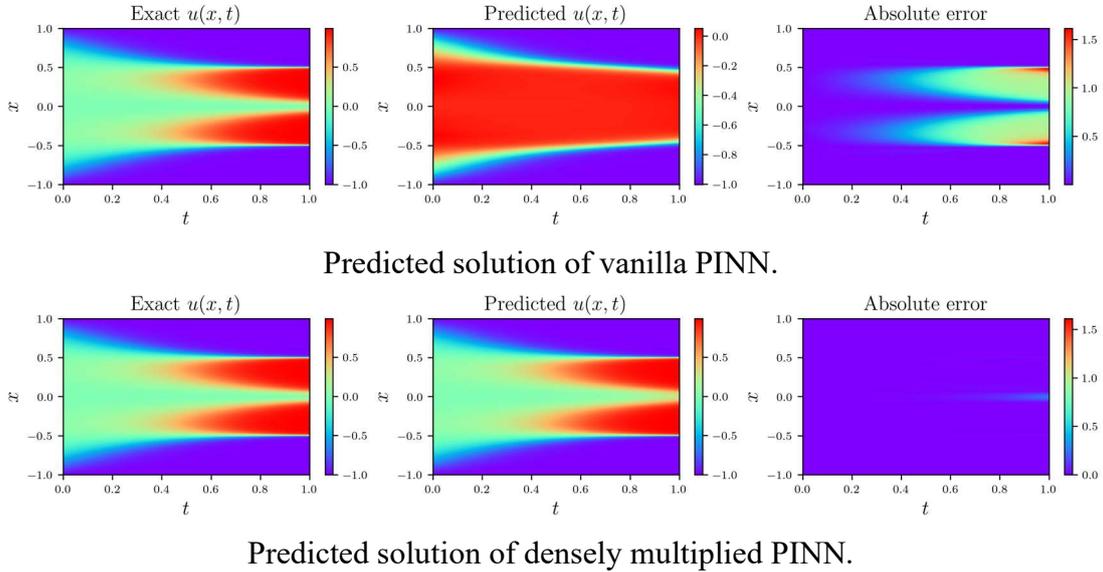

Predicted solution of vanilla PINN.

Predicted solution of densely multiplied PINN.

Figure 4: Allan-Cahn equation: Comparison between vanilla PINN and densely multiplied PINN.

Table 1: $L^2$ error comparison between different methods for the Allan-Cahn equation.

| Method | $L^2$ error | Time per iteration |
|---|---|---|
| Vanilla PINN | 5.32e-01 | 14 ms |
| ResNet | 8.00e-01 | 14 ms |
| Modified MLP | 4.31e-02 | 38 ms |
| DM-PINN | **2.57e-02** | 26 ms |

The comparison between vanilla PINN and densely multiplied PINN is shown in Fig.4. It clearly shows that our method obtains an accurate outcome, and the vanilla PINN failed to solve the equation. The further comparison with different PINN architectures is presented in Table 1. It is obvious that our method gets the highest accuracy, and it only needs 2/3 of the time used by the modified MLP model that achieves the second highest accuracy. It demonstrates the efficiency of our model as it does not introduce any extra trainable parameters.

4.2. Helmholtz equation

The Helmholtz equation is widely used in modeling wave phenomena. Here we study the equation in the form below:

$$\frac{\partial^2 u}{\partial x^2} + \frac{\partial^2 u}{\partial y^2} + k^2 u - q(x, y) = 0, \ x \in [-1, 1], \ y \in [-1, 1], \tag{15}$$

$$u(-1, y) = u(1, y) = u(x, -1) = u(x, -1) = 0, \tag{16}$$

where the source term $q(x, y)$ is in the form of:

$$\begin{aligned} q(x, y) = &-(a_1\pi)^2 \sin(a_1\pi x)\sin(a_2\pi y) \\ &-(a_2\pi)^2 \sin(a_1\pi x)\sin(a_2\pi y) \\ &+k^2 \sin(a_1\pi x)\sin(a_2\pi y) w, \end{aligned} \tag{17}$$

here we set $a_1 = 1$, $a_2 = 4$ and $k = 1$.

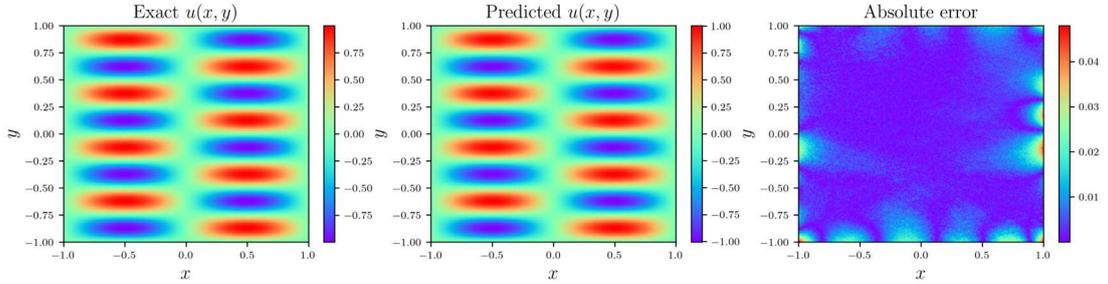

Predicted solution of vanilla PINN.

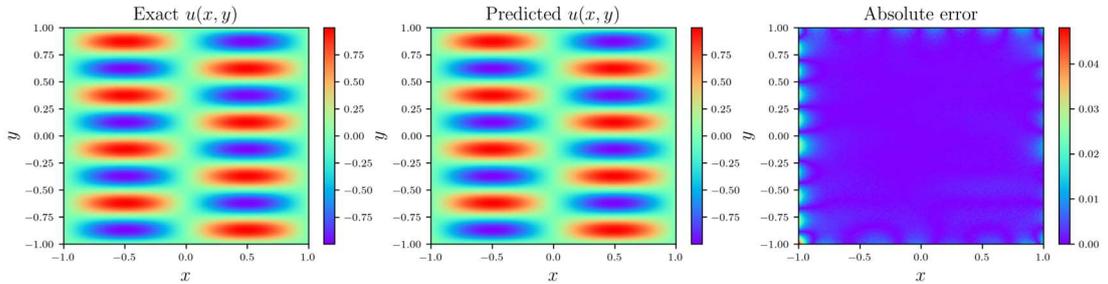

Predicted solution of densely multiplied PINN.

Figure 5: Helmholtz equation: Comparison between vanilla PINN and densely

multiplied PINN.

For this instance, we choose a neural network with 4 hidden layers, and each hidden layer has 50 neurons. The number of training points $N_r$, $N_{ic}$ and $N_{bc}$ are 20000, 200 and 200. The weight coefficients are all set as 1. The training is done with 15000 iterations, and the learning rate is set as 0.002. The results of the vanilla method and DM-PINN are shown in Fig.5. Evidently, our method obtains smaller absolute error at the boundaries. The accurate solution of the boundary condition can help to prevent the propagation failures problem [12]. It means the inner part of the whole domain can also be properly handled, which is apparent in the comparison of absolute error distributions of the two models. Table 2 shows the comparisons of different PINN architectures. Among all the models, DM-PINN obtains the minimum average $L^2$ error of 5.29e-3.

Table 2: $L^2$ error comparison between different methods for the Helmholtz equation.

| Method | $L^2$ error | Time per iteration |
|---|---|---|
| Vanilla PINN | 1.66e-2 | 27 ms |
| ResNet | 1.64e-2 | 27 ms |
| Modified MLP | 6.33e-3 | 69 ms |
| DM-PINN | **5.29e-3** | 39 ms |

4.3. Burgers' equation

The Burgers' equation used here could be expressed as:

$$u_t + uu_x - (0.01/\pi)u_{xx} = 0, \; x \in [-1, 1], \; t \in [0, 1], \quad (18)$$

$$u(0, x) = -\sin(\pi x), \quad (19)$$

$$u(t, -1) = u(t, 1) = 0. \quad (20)$$

The choice of training points in this case are $N_r = 10000$, $N_{ic} = 100$, $N_{bc} = 200$. The skip densely multiplied structure is adopted here. For this example, we chose a neural network structure containing 6 hidden layers with 80 neurons each layer. The weight coefficients are all set as 1. The learning rate is set as 0.001.

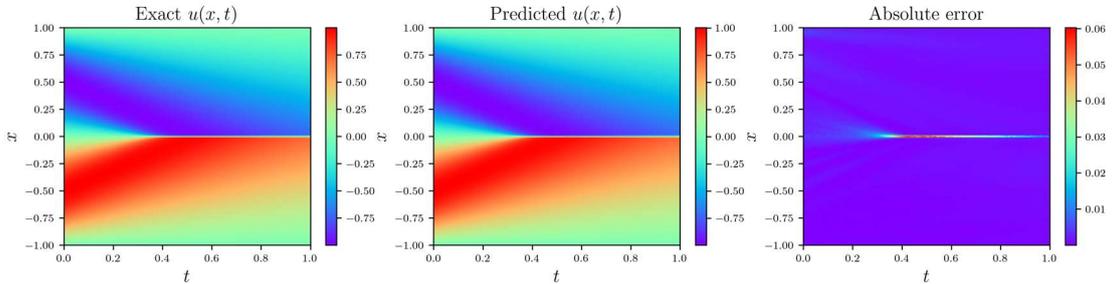

Predicted solution of vanilla PINN.

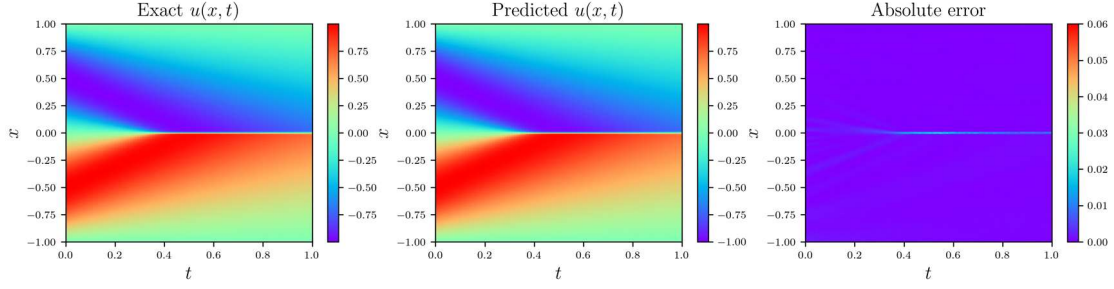

Predicted solution of skip densely multiplied PINN.

Figure 6: Burgers' equation: Comparison between vanilla PINN and skip densely multiplied PINN.

The outcomes of vanilla PINN and our method are shown in Fig.6. It can be observed from the absolute error distribution that our method obtains a smaller absolute error at the central area where the gradient of the solution is large. The $L^2$ errors and running time for 1 iteration of different PINN architectures are listed in Table 3. It shows that our method and modified MLP achieve close accuracy. However, the time cost of our method is 15 ms less than the modified MLP for one iteration, which demonstrates our model needs less computation than the modified MLP.

Table 3: $L^2$ error comparison between different methods for the Burgers' equation.

| Method | $L^2$ error | Time per iteration |
| --- | --- | --- |
| Vanilla PINN | 1.18e-2 | 20 ms |
| ResNet | 5.26e-3 | 20 ms |
| Modified MLP | 1.84e-3 | 49 ms |
| SDM-PINN | **1.79e-3** | 34 ms |

4.4.1D convection equation

The final case we studied is the 1D convection equation that models the transport phenomenon. The equation is described as follows:

$$\frac{\partial u}{\partial t} + \beta \frac{\partial u}{\partial x} = 0,\ x \in [0,\ 2\pi],\ t \in [0,\ 1], \tag{21}$$

$$u(0,\ x) = h(x), \tag{22}$$

$$u(t,\ 0) = u(t,\ 2\pi), \tag{23}$$

where $h(x)$ denotes the initial condition. Here we set $h(x) = \sin(x)$. Periodic boundary condition is adopted here. The previous works showed that it would be hard for the vanilla PINN method to solve this PDE when the convection coefficient $\beta$ gets lager [10, 12]. Thus, we set $\beta$ as 30 to examine the capacity of our method.

The skip densely multiplied structure is adopted to deal with this problem. In this example, the structure of a neural network contains 8 hidden layers with 60 neurons each layer. The choices of training points are $N_r = 20000$, $N_{ic} = 200$, $N_{bc} = 400$. The weight coefficients are all set as 1. The model was trained for 10000 iterations with Adam optimizer, and the learning rate is set as 0.001. The comparison between SDM-

PINN and vanilla PINN is shown in Fig.7. Obviously, the vanilla PINN fails to solve this problem. It succeeds in getting the correct solution in only a few time steps near the initial state, and the accuracy becomes worse as time goes on. However, our method can provide an accurate approximation. The comparison between different models is shown in Table 4. For this instance, our method achieves smaller error by over one order magnitude than all the other structures. The proposed method is the only model that successfully solves this problem.

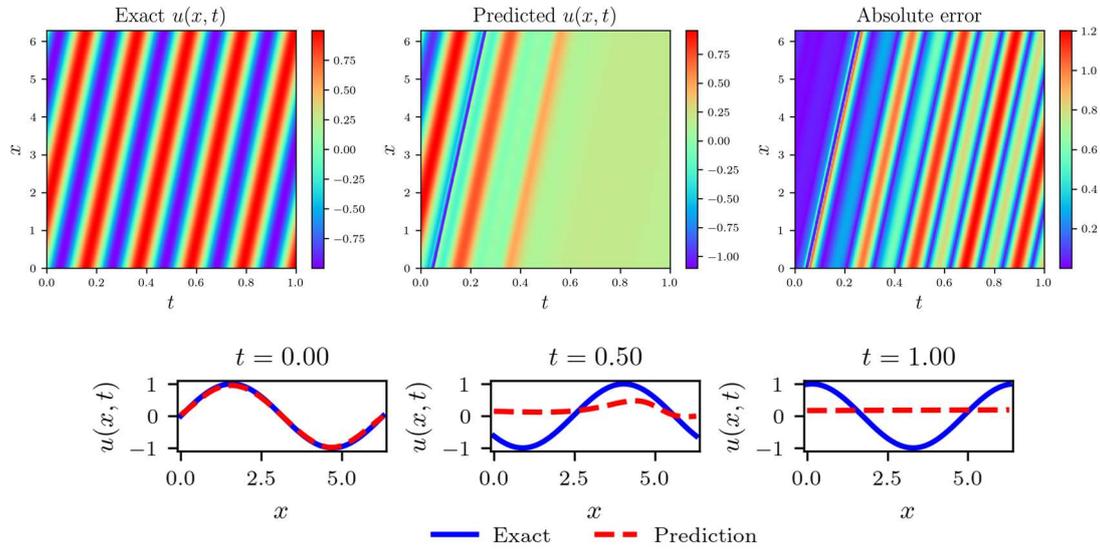

Predicted solution of vanilla PINN.

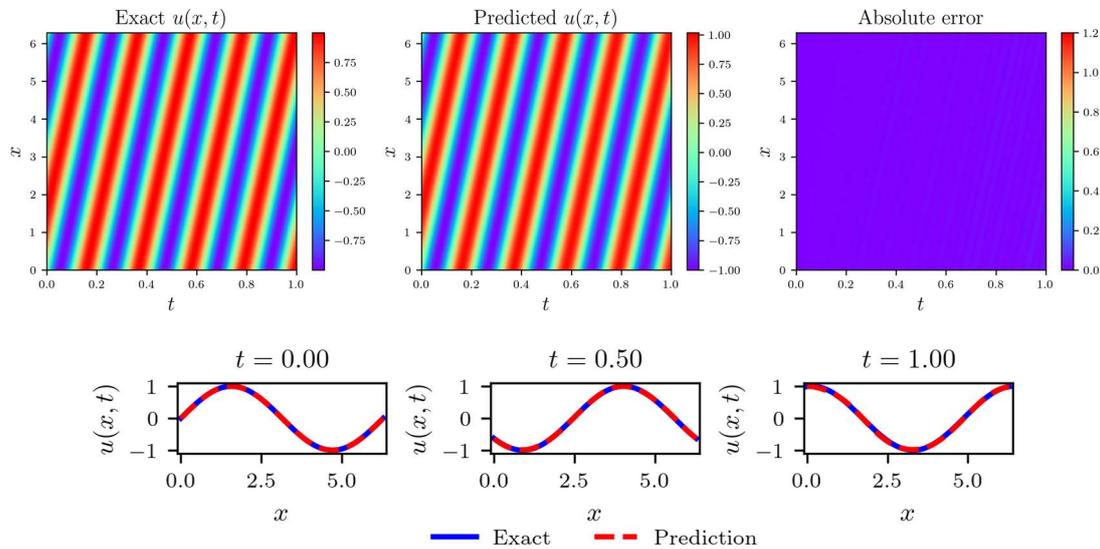

Predicted solution of skip densely multiplied PINN.

Figure 7: 1D convection equation: Comparison between vanilla PINN and skip densely multiplied PINN.

Table 4: $L^2$ error comparison between different methods for the 1D convection equation.

| Method | $L^2$ error | Time per iteration |
| --- | --- | --- |
| Vanilla PINN | 6.78e-01 | 28 ms |
| ResNet | 7.30e-01 | 29 ms |
| Modified MLP | 8.64e-01 | 65 ms |

| SDM-PINN | 1.93e-02 | 43 ms |
|---|---|---|

## 5. Discussion

To further analyse the performance of the proposed method, firstly, we compare the largest eigenvalue $\lambda_{max}$ of the Hessian matrix of the loss function $\nabla_\theta^2 \mathcal{L}(\theta)$ between each model. Previous studies have confirmed that the $\lambda_{max}$ of the Hessian matrix is highly determined by the network architecture [38], and the optimum learning rate for a model should locate between 0 and $2/\lambda_{max}$ [39]. However, in the practical training process, computing the eigenvalue of the Hessian matrix could be computationally expensive, and the optimum learning rate might be too small to make the model converge within a reasonable amount of time. Although we didn't use the optimum learning rate directed by $2/\lambda_{max}$, we can still use $\lambda_{max}$ to assess the performance of the model. Wang et al. [24] mentioned that the $\lambda_{max}$ of $\nabla_\theta^2 \mathcal{L}(\theta)$ could reflect the stiffness of the gradient flow dynamics during the training process. The larger $\lambda_{max}$ is, the stiffer the model is. Here we take the Helmholtz equation solved above as an example, the largest eigenvalues of $\nabla_\theta^2 \mathcal{L}(\theta)$ during the whole training process of each model are shown in Fig.8. Apparently, the largest eigenvalue of our model stays as the smallest one among all the models. It indicates that our model has less restriction in the required learning rate.

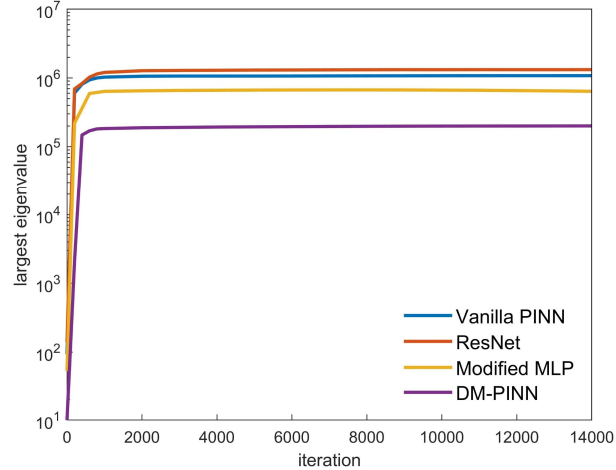

Figure 8: Largest eigenvalue of the Hessian during the training.

To show in detail, Fig.9. presents the loss and $L^2$ error curves during the training process. It can be observed from Fig.9. (a) that at the beginning of the training process, all the loss curves stay stable. With the training process going on and the loss function getting close to the minima, oscillations occur. Among the 4 models, the oscillation of our model appears to be the lightest one. It is well known that the oscillation around the minima will cause a negative influence on the convergence of the model [40]. Thus, a stable loss curve also means the training of the model could be easier. This is verified by Fig.9. (b). It shows that during the whole training process, the $L^2$ errors of both modified MLP and DM-PINN stay smaller than those of vanilla PINN and ResNet structure. It could be easily found that as the iteration increases, the $L^2$ error of DM-PINN decreases faster than that of modified MLP. We also conduct the training with different learning rates with their results shown in Table 5. All the outcomes show that our method gets the best results among the 4 models.

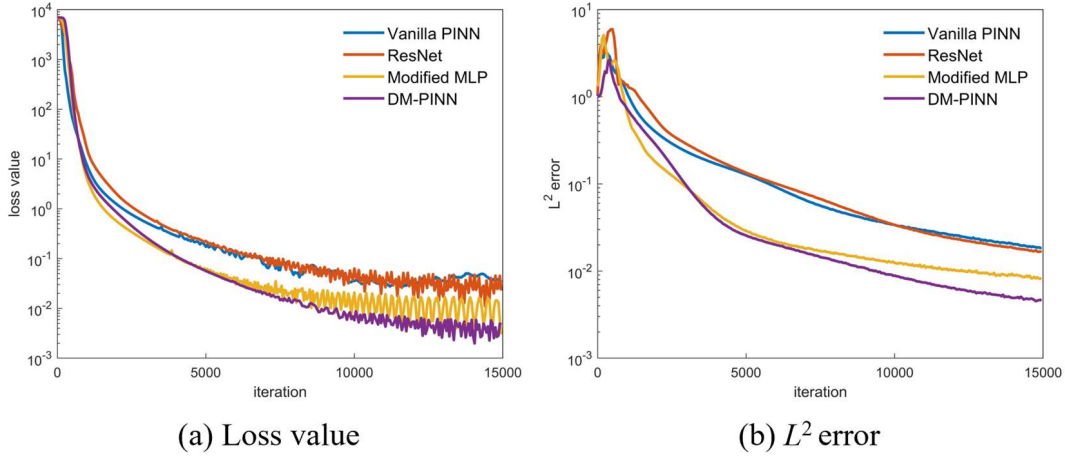

(a) Loss value　　　　　　　　　　(b) $L^2$ error

Figure 9: Loss curves and $L^2$ error curves of different models.

Table 5: $L^2$ errors corresponding to different learning rates and different models.

| Learning rate | Vanilla PINN | ResNet | Modified MLP | DM-PINN |
| --- | --- | --- | --- | --- |
| 0.001 | 1.55e-02 | 1.70e-02 | 7.06e-03 | **6.54e-03** |
| 0.005 | 1.33e-02 | 1.55e-02 | 6.40e-03 | **6.34e-03** |
| 0.01 | 1.92e-02 | 1.83e-02 | 7.48e-03 | **5.72e-03** |

Secondly, we measure the efficiency of each model. We train the 4 models for 10 minutes with 5 independent trials. The average accuracies of each model are listed in Table 6. We can observe that our method achieves the highest accuracy among all 4 cases. It indicates that our model has better efficiency than the other 3 structures.

Table 6: $L^2$ errors corresponding to different PDEs and different models.

| PDE | Vanilla PINN | ResNet | Modified MLP | Ours |
| --- | --- | --- | --- | --- |
| Allan-Cahn | 9.40e-2 | 5.04e-1 | 3.83e-2 | **1.34e-2** |
| Helmholtz | 1.11e-2 | 1.07e-2 | 1.22e-2 | **4.90e-3** |
| Burgers' | 4.72e-3 | 3.44e-3 | 1.17e-3 | **1.04e-3** |
| 1D convection | 3.00e-2 | 3.13e-2 | 8.67e-1 | **1.16e-2** |

**Conclusion**

In this paper, we proposed a novel PINN architecture with densely multiply PINN (DM-PINN). It introduces the element-wise multiplication between a hidden layer and its following hidden layers. By using this structure, the output of a hidden layer can be reused, thus the expressive power of neural network can be enhanced without increasing the number of training parameters. Experimental results on several benchmarks show that our method achieved more accurate predictions with the same training resources. Gradient flow dynamics analysis indicated that our method could contribute to lower the stiffness of neural network and stabilize the training process. Thus, we can achieve a faster decay in loss function and $L^2$ error. This approach can be used as a basic structure for the PINN method and be compatible with more advanced training processes to resolve complex problems.